\documentclass[11pt]{article}
\usepackage{acl2010}
\usepackage{times}
\usepackage{url}
\usepackage{latexsym}
\usepackage{graphicx}
\usepackage{amsmath}

\title{Semi-automatic WordNet Linking using Word Embeddings}

\author{
\textbf{Kevin Patel}\textsuperscript{$\dagger$}, \textbf{Diptesh  Kanojia}\textsuperscript{$\dagger$,$\clubsuit$,$\star$},  \textbf{Pushpak Bhattacharyya}\textsuperscript{$\dagger$} \\
\textsuperscript{$\dagger$}Indian Institute of Technology Bombay, India\\
\textsuperscript{$\clubsuit$}IITB-Monash Research Academy, India\\
\textsuperscript{$\star$}Monash University, Australia\\
\textsuperscript{$\dagger$}\{kevin.patel, diptesh, pb\}@cse.iitb.ac.in\\
}

\date{}

\begin{document}
\maketitle

\begin{abstract}

Wordnets are rich lexico-semantic resources. Linked wordnets are extensions of wordnets, which link similar concepts in wordnets of different languages. Such resources are extremely useful in many Natural Language Processing (NLP) applications, primarily those based on knowledge-based approaches. In such approaches, these resources are considered as gold standard/oracle. Thus, it is crucial that these resources hold correct information. Thereby, they are created by human experts. However, manual maintenance of such resources is a tedious and costly affair. Thus techniques that can aid the experts are desirable. In this paper, we propose an approach to link wordnets. Given a synset of the source language, the approach returns a ranked list of potential candidate synsets in the target language from which the human expert can choose the correct one(s). Our technique is able to retrieve a winner synset in the top 10 ranked list for 60\% of all synsets and 70\% of noun synsets.

\end{abstract}

\section{Introduction}

Wordnets \cite{fellbaum1998wordnet} have been useful in different Natural Language Processing applications such as Word Sense Disambiguation \cite{tufics2004fine,sinha2006approach}, Machine Translation \cite{knight1994building} \textit{etc.} 

Linked Wordnets are extensions of wordnets. In addition to language specific information captured in constituent wordnets, linked wordnets have a notion of an interlingual index, which connects similar concepts in different languages. Such linked wordnets have found their application in machine translation \cite{hovy1998combining}, cross-lingual information retrieval \cite{gonzalo1998indexing}, \textit{etc.}

Given the extensive application of wordnets in different NLP applications, maintenance of wordnets involves expert involvement. Such involvement is costly both in terms of time and resources. This is further amplified in case of linked wordnets, where experts need to have knowledge of multiple languages. Thus, techniques that can help reduce the effort needed by experts are desirable.

Recently, deep learning has been extremely successful in a wide array of NLP applications. This is primarily due to the development of word embeddings, which have become a crucial component in modern NLP. They are learned in an unsupervised manner from large amounts of raw corpora. \newcite{Bengio2003JLMR} were the first to propose neural word embeddings. Many word embedding models have been proposed since then \cite{Collobert2008ICML,Huang2012ACL,Mikolov2013NIPS,Levy2014ACL}. They have been efficiently utilized in many NLP applications: Part of Speech Tagging \cite{Collobert2008ICML}, Named Entity Recognition \cite{Collobert2008ICML}, Sentence Classification \cite{Kim2014EMNLP}, Sentiment Analysis \cite{Liu2015EMNLP}, Sarcasm Detection \cite{Joshi2016EMNLP}

\newcite{2013Mikolovtr} made a particularly interesting observation about the structure of the embedding space of different languages. They noted that there is a linear mapping between such spaces. 

\begin{figure*}[ht]
\centering
\includegraphics[width=0.85\linewidth]{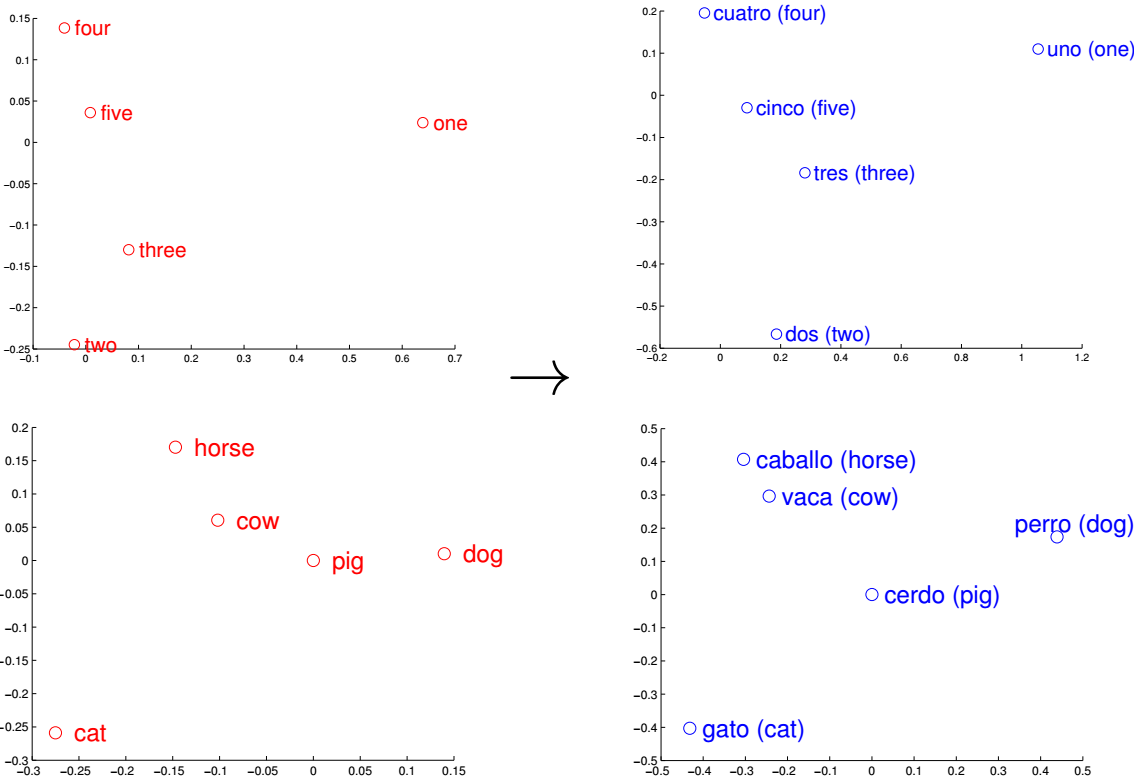}
\caption{Word embeddings of numbers and animals in English (left) and Spanish (right) (taken from \cite{2013Mikolovtr}).}
\label{fig:mikolov}
\end{figure*}

In this paper, we address the following question:
\begin{center}
	\textit{``Can information about the structure of embedding spaces of different languages and the relation among them be used to aid linking of corresponding wordnets?''}
\end{center}

We demonstrate that this is true at least in the case of English and Hindi WordNets. We propose an approach to link them using word embeddings. Given a synset of the source language, the approach provides a ranked list of target synsets. This makes the overall linking task easy for human experts, as they have to choose from a relatively small set of potential candidates. Our evaluation shows that our technique is able to retrieve a winner synset in the top 10 ranked list for 60\% and 70\% of all synsets and noun synsets respectively.


\section{Background and Related Work}

Princeton WordNet or the English WordNet was the first wordnet and inspired the development of many other wordnets. EuroWordNet \cite{1997Vossen} is a linked wordnet comprising of wordnets for European languages, \textit{viz}, Dutch, Italian, Spanish, German, French, Czech and Estonian. Each of these wordnets is structured in the same way as the Princeton WordNet for English \cite{1990Miller} - synsets (sets of synonymous words) and semantic relations between them. Each wordnet separately captures a language-specific information. In addition, the wordnets are linked to an Inter-Lingual-Index, which uses Princeton WordNet as a base. This index enables one to go from concepts in one language to similar concepts in any other language. Such features make this resource helpful in cross-lingual NLP applications.

IndoWordNet \cite{2010Bhattacharyya} is a linked wordnet comprising of wordnets for major Indian languages, \textit{viz}, Assamese, Bengali, Bodo, Gujarati, Hindi, Kannada, Kashmiri, Konkani, Malayalam, Manipuri, Marathi, Nepali, Oriya, Punjabi, Sanskrit, Tamil, Telugu, and Urdu. These wordnets have been created using the expansion approach using Hindi WordNet as a pivot, which is partially linked to English WordNet.
Previously, \newcite{joshi2012eating} come up with a heuristic based measure where they use bilingual dictionaries to link two wordnets. They combine scores using various heuristics and generate a list of potential candidates for linked synsets.

\newcite{singh2016mapping} discuss a method to improve the current status of Hindi-English linkage and present a generic methodology \textit{i.e.,} manually creating bilingual mappings for concepts which are unavailable in either of the languages or not present as a synset in the target wordnet. Their method is beneficial for culture-specific synsets, or for non-existing concepts; but, it is cost and time inefficient, and requires a lot of manual effort on the part of a lexicographer.

Our approach is mainly geared towards reducing effort on the part of the lexicographers.

\section{Problem Statement}

Given wordnets of two different languages $E$ and $F$ with sets of synsets $\{ s_E^1, s_E^2, \ldots, s_E^m\}$ and $\{ s_F^1, s_F^2, \ldots, s_F^n\}$ respectively, find mappings of the form $<s_E^i, s_F^j>$ which are semantically correct.

\section{Approach}

We adapted the technique of translating words in \newcite{2013Mikolovtr} to translate synsets (see fig \ref{fig:mikolov}). In order to do so, however, we need "synset embeddings". We computed the same by assigning to a synset-id, the average of the "word embeddings" of its synset-members. To the best of our knowledge, this is a first attempt at solving this problem using word embeddings. The following is a detailed description of the technique.

Let $E$ and $F$ be two languages. Let $|E|$ and $|F|$ be the number of synsets in wordnets of $E$ and $F$ respectively. Let $s_E^i$ and $s_F^j$ be the $i^{th}$ and $j^{th}$ synsets of $E$ and $F$ respectively, with $s_E^i =\{ e_\alpha^1, e_\alpha^2, \ldots, e_\alpha^{m_i}\}$ and  $s_F^j =\{ f_\beta^1, f_\beta^2, \ldots, f_\beta^{n_j} \}$, where $e_\alpha^p$ and $f_\beta^q$ are words in vocabulary of $E$ and $F$ respectively for $1 \leq p \leq m_i$ and $1 \leq q \leq n_j$, and $ 1 \leq i \leq |E|$ and $ 1\leq j \leq |F|$. 

Let $v_{e_\alpha^p}$ be the word embedding corresponding to $e_\alpha^p$. Then we estimate $v_{s_E^i}$, the embedding for synset $s_E^i$, as 
\begin{equation}
v_{s_E^i} = \frac{1}{m_i}\sum\limits_{p=0}^{m_i}v_{e_\alpha^p}
\end{equation}
Similarly, 
\begin{equation}
v_{s_F^j} = \frac{1}{n_j}\sum\limits_{q=0}^{n_j}v_{f_\beta^q}
\end{equation}

Given links of the form $\left\langle s_E^i , s_F^j\right\rangle$, we learn $W$ such that the error $Err$
\begin{equation}
Err = \|W.v_{s_E^i} - v_{s_F^j}\|^2
\end{equation}
is minimized.

Now, to find a mapping for a new synset $s_E^k$, one needs to
\begin{enumerate}
	\item Calculate $v' = W.v_{s_E^k}$
	\item Find $v_{s_F^l}$ such that $v_{s_F^l}.v'$ is maximized
	\item Create link $\left\langle s_E^k , s_F^l\right\rangle$
\end{enumerate}

Our hypothesis is that for a given synset-id, the noise added to its representative embedding by a highly polysemous synset-member will be canceled out, while the actual information content pertaining to that synset-id will be enhanced, due to contribution from other, relatively less polysemous, synset members.

\section{Experiments}
\subsection*{Datasets}
We applied our technique to link Hindi and English Wordnets. We obtained a dataset of mappings between English and Hindi wordnets from the developers of IndoWordNet. These mappings are of the form $\left\langle hindi\_synset\_id, english\_synset\_id, link\_type\right\rangle$, where $link\_type$ $\in$ \{DIRECT, HYPERNYMY, \textit{etc.}\}. For this experiment, we focused solely on DIRECT links. There are a total of 6,883 such mappings, the distribution among classes of which is mentioned in table \ref{table:dist}

\begin{table}[h]
\centering
\begin{tabular}{|c|c|}
\hline
\textbf{Class} & \textbf{Count} \\ \hline
Noun           & 4757           \\ \hline
Adjective      & 1283           \\ \hline
Verb           & 680            \\ \hline
Adverb         & 143            \\ \hline
\end{tabular}
\caption{Distribution of available links among various classes}
\label{table:dist}
\end{table}

For the English language, we used the pre-trained word embeddings published by Google that were trained on part of Google News Dataset (about 100 billion tokens). These embeddings are of dimension 300, and are created using CBOW model with negative sampling. 
For the Hindi language, we trained word embeddings on BOJAR HindMonoCorp dataset \cite{2014Bojar}. \newcite{2013_Mikolov_1} suggests that the input embeddings' dimension should be at least 2.5 to 4 times that of the output dimension. But we also wanted to check what happens when they are equal. Therefore, we trained two sets of embeddings, one of dimension 300, and the other of dimension 1200.

\subsection*{Evaluation Metric}
We use the accuracy@$n$ measure, i.e the prediction is said to be correct if one out of the top $n$ results returned is correct. This is because accuracy@1 is an underestimate of the system's performance, as higher-ranking synonym translations will be counted as mistakes.

\begin{figure}[hbtp]
\centering
\includegraphics[scale=0.4]{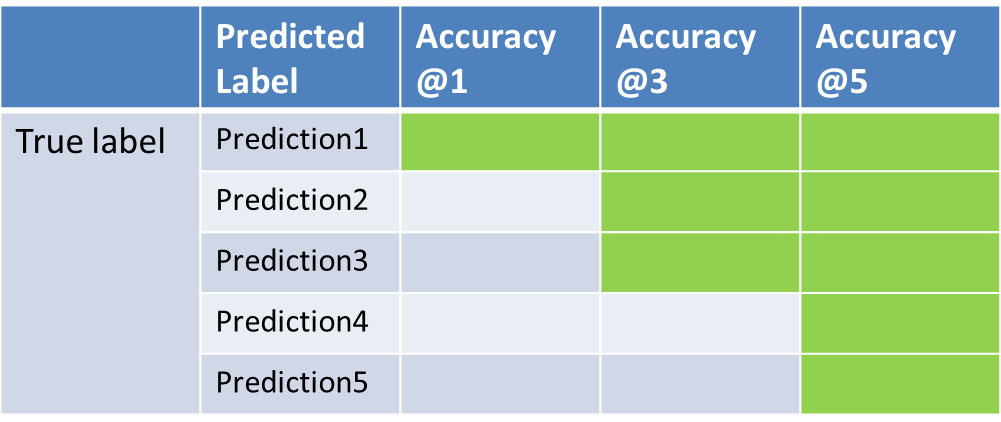}
\caption{Accuracy@n: The green colored cells indicate the predictions considered for exact match for a given accuracy@n}
\end{figure}

\section{Results and Discussion}
Table \ref{table:overall} shows the overall accuracy@n of the system, for different values of n. We also performed a per word-class evaluation, along with different settings for the embedding dimensions. Table \ref{table:class3} and Table \ref{table:class12} shows the accuracy for different word classes \footnote{All values reported are the average values obtained from 3-fold cross validation.}.

\begin{table}[hbtp]
\centering
\resizebox{0.48\textwidth}{!}{
\begin{tabular}{|c|c|c|c|c|}
\hline
\textbf{Acc@1} & \textbf{Acc@3} & \textbf{Acc@5} & \textbf{Acc@8} & \textbf{Acc@10} \\ \hline
0.29 & 0.45 & 0.52 & 0.58 & 0.60 \\
\hline
\end{tabular}
}
\caption{Results for the overall setting: Dimension of English embeddings=300, Dimensions of Hindi embeddings=300}
\label{table:overall}
\end{table}


\begin{table}[hbtp]
\centering
\resizebox{0.48\textwidth}{!}{
\begin{tabular}{|c|c|c|c|c|c|}
\hline
\textbf{Word Class} & \textbf{Acc@1} & \textbf{Acc@3} & \textbf{Acc@5} & \textbf{Acc@8} & \textbf{Acc@10} \\ \hline
Noun & 0.35 & 0.53 & 0.60 & 0.65 & 0.67 \\
Adjective & 0.26 & 0.44 & 0.50 & 0.57 & 0.60 \\
Verb & 0.15 & 0.25 & 0.29 & 0.33 & 0.37 \\
Adverb & 0.28 & 0.51 & 0.59 & 0.70 & 0.73 \\
\hline
\end{tabular}
}
\caption{Results for the setting: Dimension of English embeddings=300, Dimensions of Hindi embeddings=300}
\label{table:class3}
\end{table}


\begin{table}[hbtp]
\centering
\resizebox{0.48\textwidth}{!}{
\begin{tabular}{|c|c|c|c|c|c|}
\hline
\textbf{Word Class} & \textbf{Acc@1} & \textbf{Acc@3} & \textbf{Acc@5} & \textbf{Acc@8} & \textbf{Acc@10} \\ \hline
Noun & 0.35 & 0.52 & 0.58 & 0.63 & 0.66 \\
Adjective & 0.12 & 0.20 & 0.24 & 0.30 & 0.32 \\
Verb & 0.17 & 0.27 & 0.32 & 0.35 & 0.39 \\
Adverb & 0.38 & 0.52 & 0.65 & 0.76 & 0.80 \\
\hline
\end{tabular}
}
\caption{Results for the setting: Dimension of English embeddings=300, Dimensions of Hindi embeddings=1200}
\label{table:class12}
\end{table}


We observe that except for verbs, the approach performs decently. Here we mention some of the reasons for poor performance, as well as possible methods to address them.
\begin{itemize}
	\item The approach to create synset embeddings is inadequate. The current averaging approach only takes the synset members into account, while ignoring gloss and examples, which could provide additional information. A potential candidate approach for creating synset embeddings should properly utilize the set of French synonyms, gloss, example sentences, and synset relations.
	\item Synset members are often phrases instead of words. Creating phrase embeddings is a different problem altogether.
	\item Currently, we utilized a word embedding model which gives only one embedding per word. That is one of the reasons for ambiguity. A model which provides one embedding per sense of a word will be a more appropriate.
	\item The linear transformation approach is incorrect. While \cite{2013Mikolovtr} shows the linear relation between English and Spanish languages, this may not be true for all pairs of languages. 
	\item Perhaps, something is fundamentally missing in word embeddings. Probably presence of only co-occurrence information and lack of other information such as word order, argument frames( for verbs), \textit{etc.} leads to this poor performance.
\end{itemize}
However, we were unable to find an explanation for the degradation of results of adjectives when using 1200 dimensions for Hindi word embeddings.

\section{Conclusion and Future Work}

In this paper, we described an approach to link wordnets. It entails creating synset embeddings using the word embeddings of the synset members, and learning a function to map the embedding of a synset from the source language to an embedding in the space of target language, and returning the nearest neighbors as potential candidates for linking. Our evaluation shows that our technique is able to retrieve a winner synset in the top 10 ranked list for 60\% and 70\% of all synsets and noun synsets, respectively. Although, it did not achieve significantly good results for other classes, especially verbs. We discussed the possible reasons for poor performance and suggested mechanisms to address the same.

In future, we plan to continue this work, and explore each of the above possible reasons for poor performance, in order to mitigate them. We will also evaluate it in an active learning setting. Eventually, we aim to integrate our work with tools such as the ones created by \newcite{2012PB}, \textit{etc.} so that our work can be used by lexicographers and researchers alike.

\bibliographystyle{acl}
\bibliography{acl}
\end{document}